# Candidate Constrained CRFs for Loss-Aware Structured Prediction

Faruk Ahmed
Virginia Tech
faruk@vt.edu

Daniel Tarlow
Microsoft Research
dtarlow@microsoft.com

Dhruv Batra
Virginia Tech
dbatra@vt.edu

## Abstract

*When evaluating computer vision systems, we are often concerned with performance on a task-specific evaluation measure such as the Intersection-Over-Union score used in the PASCAL VOC image segmentation challenge. Ideally, our systems would be tuned specifically to these evaluation measures. However, despite much work on loss-aware structured prediction, top performing systems do not use these techniques. In this work, we seek to address this problem, incorporating loss-aware prediction in a manner that is amenable to the approaches taken by top performing systems. Our main idea is to simultaneously leverage two systems: a highly tuned pipeline system as is found on top of leaderboards, and a traditional CRF. We show how to combine high quality candidate solutions from the pipeline with the probabilistic approach of the CRF that is amenable to loss-aware prediction. The result is that we can use loss-aware prediction methodology to improve performance of the highly tuned pipeline system.*

## 1. Introduction

A key challenge in computer vision is to develop systems that perform well according to meaningful evaluation metrics. While there is much theory on building loss-aware systems based on Empirical Risk Minimization (ERM) and Bayesian Decision Theory (BDT), this theory is notably absent in systems that perform competitively on popular benchmarks like the PASCAL Segmentation challenge [5], such as O2P [3] or a deep-learning based pipeline [7].

Why is this the case? It is clearly not due to a lack of ideas about how to perform loss-aware learning. Consider, for example, the Intersection-Over-Union (IOU) metric that is used in the PASCAL challenge. Existing work shows how to train Structural SVMs under the IOU score [15, 18], maximize expected IOU score [9], and how to employ BDT with IOU scores [13, 16]. However, none of these ideas seem to be predominant in state-of-the-art systems that are evaluated under this metric.

We suspect that the primary issue preventing application of loss-aware learning on real problems is a mismatch in basic assumptions: the ERM and BDT-based approaches assume that there is a single probabilistic model or energy function that encodes all the modeling assumptions, and that one only needs to tune parameters of this model. On the other hand, modern state-of-art systems [3,7] are often pipeline procedures, consisting of a sequence of operations based on intuitions developed by experienced domain experts, without a single energy function or a single probabilistic model. For instance, in PASCAL, most state-of-art pipelines involve extracting category-independent segment proposals, which are then scored according to hand-designed [3] or learned features [7], followed by a greedy "pasting" procedure to form a semantic segmentation.

**Goal and Overview.** Our aim in this work is to bring together recent ideas about loss-aware learning with the pipeline systems that perform best on current benchmarks. We are inspired by recent work [14] which introduced an ERM based loss-aware predictor by minimizing expected loss over a small set of candidate solutions. Our key novelty over [14] is that we do not restrict attention to a small number of candidate solutions when approximating expected loss but instead build a Conditional Random Field (CRF) that uses the candidate solutions to focus probability mass around regions of interest. As a result, we can leverage the high quality solutions from the pipeline procedures, and additionally have access to all of the more sophisticated loss-aware machinery that accompanies the CRF formulation.

In Section 3, we derive our loss-aware predictors for a special class of loss functions. Implementing the predictors leads to computational challenges – we need to compute marginal probabilities and partition functions of CRFs where configurations are disallowed outside a Hamming ball. We address this in Section 4. Finally, in Section 6, we show experimentally that our new approach leads to improved performance over natural baselines on the PASCAL VOC image segmentation task.

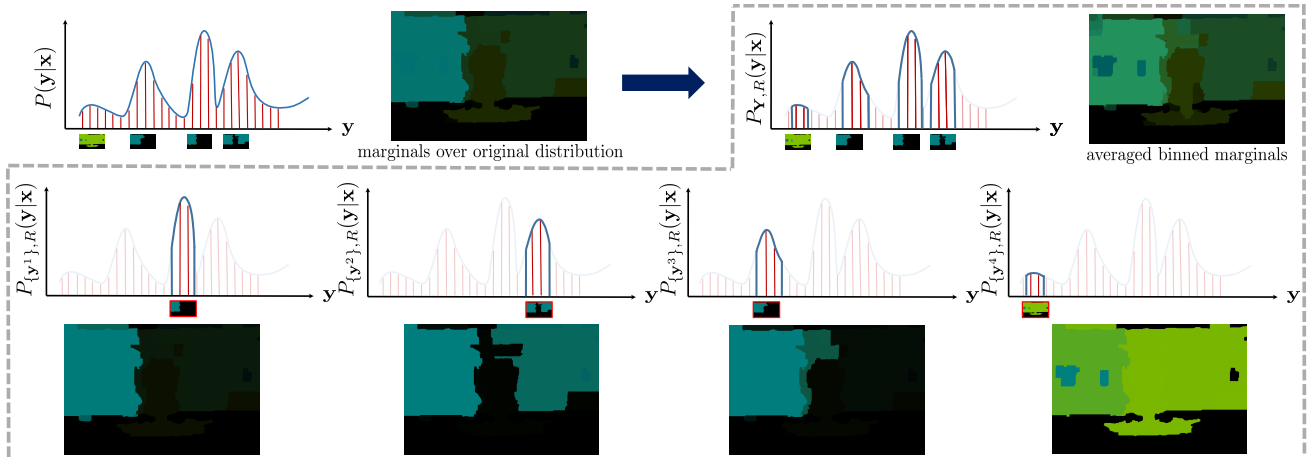

Figure 1: We use a high order potential to clamp the discrete distribution $P(\mathbf{y}|\mathbf{x})$ to Hamming balls of radius $R$ around candidate solutions $\mathbf{Y} = \{\mathbf{y}^1, ..., \mathbf{y}^4\}$, resulting in the binned distribution, $P_{\mathbf{Y},R}$. We approximate expected loss by summing expected losses in the bins. Estimating the partition function in a clamped distribution gives us the mass of a solution(bottom row), and message passing also gives us the node marginals of the clamped distributions. Both these quantities are used in our predictors.

## 2. Background

We begin by establishing our notation, and reviewing background on probabilistic structured prediction.

**Basic Notation.** For any positive integer $n$, let $[n]$ be shorthand for the set $\{1, 2, \ldots, n\}$. Given an input image $\mathbf{x} \in \mathcal{X}$, our goal is to make a prediction about output variables $\mathbf{y} \in \mathcal{Y}$, where $\mathbf{y}$ may be a figure-ground segmentation, or a category-level semantic segmentation. Specifically, let $\mathbf{y} = \{y_1 \ldots y_n\}$ be a set of discrete random variables, each taking values in a finite label set, $y_i \in Y_i$. In semantic segmentation, $i$ indexes over the (super-)pixels in the image, and these variables are the labels assigned to each (super-)pixel, i.e. $Y_i = \{\text{cat, boat, cow}, \ldots\}$. For a set $F \subseteq [n]$, we use $y_F$ to denote the tuple $\{y_i \mid i \in F\}$, and $Y_F$ to be the cartesian product of the individual label spaces $\times_{i \in F} Y_i$.

### 2.1. Conditional Random Fields and Factor Graphs

Conditional Random Fields (CRFs) are probabilistic models that represent conditional probabilities $P(\mathbf{y}|\mathbf{x})$ in a compact manner via factor graphs. Let $G = (\mathcal{V}, \mathcal{E})$ be a bipartite graph with two kinds of vertices – variable $i \in [n]$ and factor nodes $F \subseteq [n]$. Each factor holds a local compatibility function, measuring the score of the variables in its scope: $\theta_F : Y_F \to \mathbb{R}$. An edge $\{i, F\} \in \mathcal{E}$ indicates that variable $y_i$ participates in the factor function $\theta_F(\cdot)$, i.e., $i \in F$.

The score for any configuration $\mathbf{y}$ is given by $S^{\text{CRF}}(\mathbf{y}|\mathbf{x}) = \sum_F \theta_F(y_F)$, and the corresponding probability is given by the Gibbs distribution: $P(\mathbf{y}|\mathbf{x}) = \frac{1}{\mathcal{Z}} \exp S^{\text{CRF}}(\mathbf{y}|\mathbf{x})$, where $\mathcal{Z}$ is the partition function or the normalization constant.

These factor functions are typically derived from a weighted combination of features i.e., $\theta_F(y_F) = w_F^\mathsf{T} \phi(\mathbf{x}, y_F)$. From the probabilistic viewpoint, the aim is to choose parameters $\mathbf{w} = \{w_F\}$ so that $P(\mathbf{y}|\mathbf{x}; \mathbf{w})$ reflects our beliefs about how likely the different configurations of $\mathbf{y}$ are. If we wish to make a single prediction from our model, we must employ a predictor, which converts $P(\mathbf{y}|\mathbf{x}; \mathbf{w})$ into a prediction $\hat{\mathbf{y}}$. In subsequent discussion, we shall assume the dependence on $\mathbf{w}$ and not explicitly mention it.

### 2.2. Bayesian Decision Theory

Bayesian Decision Theory provides a recipe for decision making based on the principle of Maximum Expected Utility (or Minimum Expected Loss). Given a loss function $\ell(\cdot, \cdot)$ that measures of the deviation of a prediction from the ground truth, the optimal prediction is

$$\mathbf{y}^{\text{MBR}} = \operatorname*{argmin}_{\hat{\mathbf{y}} \in \mathcal{Y}} \sum_{\mathbf{y} \in \mathcal{Y}} \ell(\mathbf{y}, \hat{\mathbf{y}}) P(\mathbf{y}|\mathbf{x}). \tag{1}$$

The function being optimized over is known as the *expected loss* or *Bayes Risk*. Unfortunately, in the structured output spaces that are common in computer vision, this *Minimum Bayes Risk (MBR)* predictor is generally intractable because the minimization and summation operations are over an exponentially large output space.

### 2.3. EMBR

One recent attempt in tackling the intractability of the MBR predictor is [14], who approximated the MBR predictor by simply restricting the solution space to a small set of candidate solutions, and learning the parameters of this simplified system through ERM. Specifically, given a set of $M$ candidate solutions $\mathbf{Y} = \{\mathbf{y}^1, ..., \mathbf{y}^M\}$, a loss function $\ell(.,.)$, and the Gibbs distribution $P(.)$ corresponding to the scoring function (which may not be calibrated or reflect true beliefs), they proposed an Empirical MBR (or EMBR) predictor:

$$\mathbf{y}^{\texttt{Delta}} = \operatorname*{argmin}_{\hat{\mathbf{y}} \in \mathbf{Y}} \sum_{\mathbf{y} \in \mathbf{Y}} \ell(\mathbf{y}, \hat{\mathbf{y}}) P(\mathbf{y}|\mathbf{x}) \tag{2}$$

In this paper, we refer to the predictor as `Delta` (because of the way the distribution is assumed to be summarized by a set of delta functions at the candidate solutions). We will use these ideas of minimizing expected loss for a reduced set of solutions, and learning loss-aware predictor parameters via ERM in our approach.

### 2.4. Factorized Expected Loss Approximations

We consider better approximations for expected loss. A common case of interest is when the loss function is a sum of local loss functions, and marginal probabilities of $P(\mathbf{y}|\mathbf{x})$ can be computed efficiently. For example, if the loss function decomposes as a sum of per-variable loss functions, we have

$$\sum_{\mathbf{y} \in \mathcal{Y}} \ell(\mathbf{y}, \hat{\mathbf{y}}) P(\mathbf{y}|\mathbf{x}) = \sum_{\mathbf{y} \in \mathcal{Y}} \Big( \sum_i \ell(y_i, \hat{y}_i) \Big) P(\mathbf{y}|\mathbf{x})$$
$$= \sum_i \sum_{y_i} \ell(y_i, \hat{y}_i) P_i(y_i|\mathbf{x}), \tag{3}$$

where $P_i(y_i|\mathbf{x}; \mathbf{w})$ is the marginal probability of $y_i$. Loss functions like the Hamming loss satisfy this criteria, and thus MBR prediction is no more difficult computationally than computing marginals.

In some cases where the loss function does not decompose on variables, it turns out that the marginals can still be quite useful in order to perform MBR prediction. [13] shows that the IOU loss from the PASCAL VOC challenge is one of these cases – there is an accurate approximation to the expected loss that depends on $P(\mathbf{y}|\mathbf{x})$ only via the set of marginal probabilities

$$\mathcal{P} = \{P_i(y_i|\mathbf{x})\}_{i,y_i} = \left\{ \sum_{\hat{\mathbf{y}}:\hat{\mathbf{y}}_i = y_i} P(\hat{\mathbf{y}}|\mathbf{x}) \right\}_{i,y_i}.$$

These tractable special cases of MBR lead us to define the term *Factorized Expected Loss Approximation (FELA)*, by which we shall refer to approximations where expected loss can be expressed as a function of node marginals.

$$\sum_{\mathbf{y} \in \mathcal{Y}} \ell(\mathbf{y}, \hat{\mathbf{y}}) P(\mathbf{y}|\mathbf{x}) = f(\mathcal{P}, \hat{\mathbf{y}}) \tag{4}$$

Our goal is to make use of the structure of FELAs to do loss-aware prediction, and to leverage top-performing heuristics-based pipeline procedures.

## 3. Approach

**Overview.** We shall build structured prediction systems that combine two types of models. The first is a prediction system that we can treat as a black box, like the image segmentation pipelines discussed in the introduction; we will refer to these systems as *prediction pipelines*. From the prediction pipeline we assume access to the same interface employed by [14]: the ability to compute a score function $S^{\text{PP}}(\mathbf{y}|\mathbf{x})$ for any configuration $\mathbf{y}$, and the ability to generate a set of $M$ plausible candidate solutions using, e.g., the DivMBest approach of [20]. The second is a trained CRF that gives us the distribution $P(\mathbf{y}|\mathbf{x}) \propto \exp S^{\text{CRF}}(\mathbf{y}|\mathbf{x})$. In our experiments, we allow the CRF to be rather crudely trained— in one case we will use an SVM to train unary potentials then set pairwise potentials by hand; in the other case we will construct the CRF manually by trying to make $S^{\text{CRF}}(\mathbf{y}|\mathbf{x})$ agree with $S^{\text{PP}}(\mathbf{y}|\mathbf{x})$ on the most likely configuration and match scores for a small set of high scoring configurations. We will demonstrate that even a crudely trained CRF can be used to improve performance of the prediction pipeline by enabling better loss-aware prediction.

**Candidate Constrained CRF.** We assume that the loss function of interest admits a good FELA, so we can focus attention on producing good marginal probabilities. Our approach is to use candidate solutions produced by the prediction pipeline to restrict the probability distribution represented by the CRF. Specifically, let $\Delta^{\text{Ham}}(\mathbf{y},\mathbf{y}') = \sum_i \mathbb{1}\{y_i \neq y'_i\}$ be the Hamming distance function, and let $P_{\{\mathbf{c}\},R}(\mathbf{y}|\mathbf{x}) \propto \mathbb{1}\{\Delta^{\text{Ham}}(\mathbf{y},\mathbf{c}) \leq R\} \exp S^{\text{CRF}}(\mathbf{y}|\mathbf{x})$ be the CRF distribution constrained to lie within a Hamming ball of radius $R$ centered at the solution $\mathbf{c}$.

Recalling that $\mathbf{Y} = \{\mathbf{y}^1,\ldots,\mathbf{y}^M\}$ are the set of candidate solutions extracted from the prediction pipeline, we define the *Candidate Constrained CRF ($C^3RF$)*, which is parameterized by a radius $R$ and candidate set $\mathbf{Y}$, to be

$$P_{\mathbf{Y},R}(\mathbf{y}|\mathbf{x}) \propto \sum_{\mathbf{c}\in\mathbf{Y}} \mathbb{1}\{\Delta^{\text{Ham}}(\mathbf{y},\mathbf{c}) \leq R\} \exp S^{\text{CRF}}(\mathbf{y}|\mathbf{x}). \tag{5}$$

which can be rewritten as follows:

$$P_{\mathbf{Y},R}(\mathbf{y}|\mathbf{x}) \propto \sum_{\mathbf{c}\in\mathbf{Y}} \mathcal{Z}(\{\mathbf{c}\},R) P_{\{\mathbf{c}\},R}(\mathbf{y}|\mathbf{x}), \tag{6}$$

where $\mathcal{Z}(\mathbf{Y},R) = \sum_{\mathbf{y}\in\mathcal{Y}} \sum_{\mathbf{c}\in\mathbf{Y}} \mathbb{1}\{\Delta^{\text{Ham}}(\mathbf{y},\mathbf{c}) \leq R\} \exp S^{\text{CRF}}(\mathbf{y}|\mathbf{x})$ is the partition function of the CRF distribution that is constrained to be within Hamming radius $R$ of some configuration $\mathbf{c} \in \mathbf{Y}$. Written this way, it becomes clear that we can decompose the inference problem into $M$ separate calls, one per candidate solution. From each inference call, we will need to compute Hamming-ball constrained marginal probabilities $\mathcal{P}_{\{\mathbf{c}\},R} = \{P_{i\{\mathbf{c}\},R}(y_i|\mathbf{x})\}_{i,y_i} = \left\{\sum_{\hat{\mathbf{y}}:\hat{y}_i=y_i} P_{\{\mathbf{c}\},R}(\hat{\mathbf{y}}|\mathbf{x})\right\}_{i,y_i}$, and $\mathcal{Z}(\{\mathbf{c}\},R)$, which we will call the *mass* of solution $\mathbf{c}$. Efficient computation of these quantities will be addressed in Section 4. Note that if a configuration belongs to two Hamming balls, it will be counted twice.

**Making Predictions.** Finally, we need to use inference results in order to make loss-aware predictions. Following [14], we will restrict predictions to be one of the candidate solutions produced by the prediction pipeline. This avoids the potentially difficult argmin operation from Eq. 1. Instead, we simply predict the candidate solution $\mathbf{y} \in \mathbf{Y}$ with the minimum approximate expected loss.

Assuming a FELA exists for $\ell(.,.)$, the expected losses are functions of the marginals and the candidate solutions. So formally, our predictor can be expressed as:

$$\begin{aligned}
\mathbf{y}_{\mathbf{Y},R}^{\text{C}^3\text{RF+FELA}} &= \operatorname*{argmin}_{\hat{\mathbf{y}}\in\mathbf{Y}} \sum_{\mathbf{y}'\in\mathcal{Y}} P_{\mathbf{Y},R}(\mathbf{y}'|\mathbf{x})\ell(\mathbf{y}',\hat{\mathbf{y}}) \\
&= \operatorname*{argmin}_{\hat{\mathbf{y}}\in\mathbf{Y}} \sum_{\mathbf{c}\in\mathbf{Y}} \sum_{\mathbf{y}'\in\mathcal{Y}} \mathcal{Z}(\{\mathbf{c}\},R) P_{\{\mathbf{c}\},R}(\mathbf{y}'|\mathbf{x})\ell(\mathbf{y}',\hat{\mathbf{y}}) \\
&= \operatorname*{argmin}_{\hat{\mathbf{y}}\in\mathbf{Y}} \sum_{\mathbf{c}\in\mathbf{Y}} \mathcal{Z}(\{\mathbf{c}\},R) f(\mathcal{P}_{\{\mathbf{c}\},R},\hat{\mathbf{y}}).
\end{aligned} \tag{7}$$

We shall refer to this predictor as `C³RF+FELA`.

Note that we could also use the FELA to approximate expected loss over the unconstrained CRF. Then the predictor would take the form $\mathbf{y}_{\mathbf{Y}}^{\text{CRF+FELA}}$:

$$= \operatorname*{argmin}_{\hat{\mathbf{y}}\in\mathbf{Y}} \sum_{\mathbf{y}'\in\mathcal{Y}} P(\mathbf{y}|\mathbf{x})\ell(\mathbf{y}',\hat{\mathbf{y}}) = \operatorname*{argmin}_{\hat{\mathbf{y}}\in\mathbf{Y}} f(\mathcal{P},\hat{\mathbf{y}}) \tag{8}$$

We shall refer to this predictor as CRF+FELA.

**Solution Proposals.** In order to generate candidate solutions, we make the same assumptions about the prediction pipelineas was made in [14], namely that per-variable unary potentials can be modified, and then the perturbed minimization can be done efficiently. Like [14], we make use of the DivMBest algorithm.

DivMBest finds diverse $M$-best solutions sequentially. At each step, the next best solution is found simply by performing MAP inference on the scoring function augmented with diversity promoting factors that encourage successive solutions to be different from the previous one. With Hamming diversity, $\Delta^{\text{Ham}}(\mathbf{y}, \mathbf{y}^m)$, the $\Delta$-augmented scoring function for the $M$-th diverse solution can be written as:

$$S_\Delta^{\text{PP}}(\mathbf{y}; M) = S^{\text{PP}}(\mathbf{y}) + \sum_{m=1}^{M-1} \lambda \Delta^{\text{Ham}}(\mathbf{y}, \mathbf{y}^m) \tag{9}$$

Here, $\lambda$ acts as a *diversity* controlling parameter, with higher values encouraging more diversity.

**Learning predictor parameters.** We learn parameters that influence the generation of the set of candidate solutions - for DivMBest, this is the $\lambda$ parameter above - and $R$, the radius of the Hamming balls. Additionally, there is a temperature parameter implicit in the Gibbs distribution associated with the C$^3$RF, which we have not explicitly mentioned when not relevant to the discussion.

The parameters can be learned either through ERM on the task loss in the spirit of EMBR (Section 2.3), or in a fully BDT setting by Maximum Likelihood Estimation (MLE).

**Recap.** In summary, a broad overview of our process is: We first acquire diverse solution proposals from a high-performance prediction pipeline. Then we construct a CRF (see experiments for details on how we construct the CRFs), and compute masses and Hamming-constrained marginals on this CRF. Using these quantities, and an existing FELA, we predict the solution with minimum approximate expected loss from our set of proposals.

Next, we show how we can compute the quantities required for our FELA-based MBR predictors: the *mass* of candidate solutions and Hamming ball-constrained marginals.

## 4. Estimating masses and marginals in bins

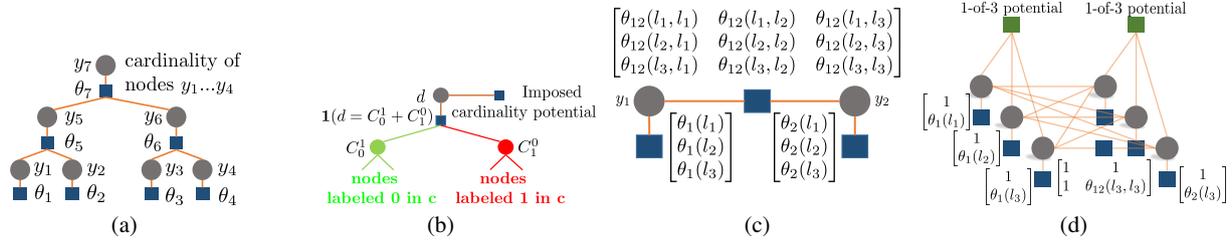

Figure 2: (a) The state of the root node ($y_7$) is the cardinality of all the leaf nodes; (b) The root node $d$ counts the Hamming distance from **c**; (c) A simple 2 node multi-label graph (each node takes 3 labels $\{l_1, l_2, l_3\}$); (d) 1-of-$K$ potentials are imposed binary nodes corresponding to the same multi-label variable to ensure that exactly one of them is ON. To avoid clutter, not all factors have not been shown.

Here we describe how we estimate *mass* of a candidate solution along with the Hamming-constrained given a factor graph, the candidate solution **c**, and a given bin radius $R$.

To the factor graph, we add a higher-order potential (HOP), contributing a score $\theta_{\text{HOP}}$, that clamps probabilities of all solutions lying outside the Hamming ball of radius $R$ to 0. Thus, the probability of a solution $\mathbf{y}$ in the constrained distribution takes the form

$$P_{\{\mathbf{c}\}, R}(\mathbf{y}) \propto \exp\left\{\sum_F \theta_F(y_F) + \theta_{\text{HOP}}(\mathbf{y}; \mathbf{c}, R)\right\}, \tag{10}$$

where the HOP, $\theta_{\text{HOP}}$, is defined as follows:

$$\theta_{\text{HOP}}(\mathbf{y}; \mathbf{c}, R) = \begin{cases} -\infty, & \text{if } \Delta^{\text{Ham}}(\mathbf{y}, \mathbf{c}) > R \\ 0, & \text{otherwise.} \end{cases} \tag{11}$$

## 4.1. Imposing Hamming constraints with the HOP

Since Hamming distance from a solution, **c**, is just the number of nodes disagreeing with **c**, we make use of cardinality potentials [6] to impose Hamming constraints in our model.

In [17], an efficient method for imposing arbitrary cardinality potentials in a sum-product message passing setting was introduced. A binary tree is constructed on top of nodes on which some arbitrary cardinality potential is intended to be imposed. Intermediate nodes in the tree compute the sum of cardinalities of their children, such that at the root node, we can compute the beliefs for the number of leaf nodes that are ON. (See Fig. 2a) An arbitrary cardinality potential can be imposed on the root node, and passing messages down to the leaves will give us revised beliefs.

**Hamming distance in binary graphs** For graphs with nodes taking binary labels, the Hamming distance between solutions **y** and **c** is the sum of the count of OFF bits in **y** where there are ON bits in **c** ($C_0^1$) and the count of ON bits in **y** where there are OFF bits in **c** ($C_1^0$). As shown in Fig. 2b, we impose the required cardinality potential, which disallows solutions outside the Hamming radius, by constructing the binary tree such that the two subtrees beneath the root node compute the two counts $C_0^1$ and $C_1^0$, and the root node sums these counts. Thus, the root node gives us the total count of disagreements between **y** and **c**. If we now set the cardinality potential to 0 when the count is $\leq R$, and $-\infty$ otherwise, we can impose a hard constraint that only allows **y** within radius $R$ of **c**.

**Hamming distance in multilabel graphs.** We first expand the multi-label graph into an equivalent binary graph. A node taking $K$ labels in the multi-label graph corresponds to $K$ nodes in the binary graph. For every such expanded set of nodes, we impose a 1-of-$K$ potential that forces only one of the nodes to be ON. We construct such a potential by once again using the cardinality tree from [17] such that the only allowed cardinality for any subset of $K$ nodes is 1.

For computing Hamming distance from a solution **c** in this model, we only need to compute the number of OFF nodes among the set of nodes that are ON in **c**, since that gives us the number of disagreements between the current state of the graph and **c**. In a similar fashion as with the binary case, we impose a Hamming constraint that disallows solutions outside the Hamming ball centered at **c**.

**Hamming constrained marginals and masses:** The partition function of this constrained CRF is the *mass* of the solution, **c**. To estimate the partition function, we first perform sum-product message passing on the HOP-augmented (expanded) factor graph thus obtaining node and factor marginals. Let $N(i)$ be the number of factors with an edge to node $i$, $\mu_i(y_i)$ be the node marginal of node $i$, $\mu_F(y_F)$ be the factor marginal for factor $F$, and $\theta_F(y_F)$ be the potential for factor $F$. Then the standard Bethe estimator for the partition function [21] is given as follows:

$$\log \mathcal{Z} = \sum_{i \in \mathcal{V}} (N(i) - 1) \left[ \sum_{y_i \in Y_i} \mu_i(y_i) \log \mu_i(y_i) \right] - \sum_{F \in \mathcal{F}} \sum_{y_F \in \mathbf{Y}_F} \mu_F(y_F) \left( \log \theta_F(y_F) + \log \mu_F(y_F) \right) \quad (12)$$

For tree-structured graphs, sum-product message passing provides exact marginals, and the Bethe approximation is exact. In loopy graphs, results are often good; in the experiments section we compare estimates returned by the Bethe estimator to a sampling-based baseline on small problems and find the Bethe estimator to be efficient in terms of accuracy.

The set of factors in the graph involve the factors in the cardinality tree. The largest factor in the graph (assuming #labels < #superpixels) would belong to the cardinality tree, and would be of $O(n^2)$ size (where $n$ is the number of nodes/superpixels), so the cost of computing the *mass* using the Bethe approximation is $O(n^2)$.

## 5. Related Work

There is significant work both inside and outside of computer vision on loss-aware decision making, some of which were reviewed in the introduction. A key difference in our approach versus many of the other loss-aware learning methods is in the use of the prediction pipeline method in order to generate candidate solutions. However, there are some works that operate within this same set of assumptions. As mentioned in Section 2.3, [14] extracts a set of candidate solutions from a prediction pipeline, and the approach in that work can be viewed as a special case of our approach that arises when the Hamming radius in the C$^3$RF is set to 0 (i.e., the model is a mixture of delta distributions at each of the candidates). We will show in the experiments that the richer set of distributions that arise from the C$^3$RF model allow us to improve over this approach.

Other methods that use candidate solutions followed by loss-aware decision making have been explored in the Machine Translation (MT) community. MBR decoding for MT was first introduced in [10], where it was used to predict the best translation from a set of $N$-best candidate translations. Subsequent work [19] [11] showed efficient ways to perform MBR prediction from a larger pool of candidate solutions in the statistical MT setting.

The other relevant line of work comes from variational inference techniques for mixture models [2,8,12]. There are two main differences in our approach. First, rather than learn the mixture components, we simply take the outputs of the prediction pipeline and use them as the centers of the mixture components. This simplifies the inference task. Second, we make use of hard mixture assignments, but we note that other softer choices could potentially be used instead, and that is a topic for future exploration.

# 6. Experiments

First, we perform experiments on synthetic data to confirm that our estimated masses are a good approximation. Then we evaluate our proposed predictors on two real datasets: Binary (foreground-background) segmentation and category level segmentation on PASCAL VOC 2012 `val` set [5].

**Baselines:** We compare our predictors to `Delta`, which is the natural baseline for EMBR methods. We also compare to MAP, which shows how much we can improve upon existing models by adding our loss-aware prediction stage.

Since the mass of a solution represents its local spread of probability, it can be more indicative of the robustness of a solution. So one simple way to generalize `Delta` is to simply replace the probabilities of the candidate solutions by the masses of the solutions.

$$\mathbf{y}^{\texttt{Mass}} = \underset{\hat{\mathbf{y}} \in \mathbf{Y}}{\operatorname{argmin}} \sum_{\mathbf{y} \in \mathbf{Y}} \ell(\mathbf{y}, \hat{\mathbf{y}}) \mathcal{Z}(\{\mathbf{y}\}, R) \tag{13}$$

We include the `Mass` predictor in our experiments as a cruder approximation of expected loss.

**FELA:** Our experiments on segmentation are evaluated with the Intersection-Over-Union(IOU) loss function. In [13] and [16], a FELA was presented for IOU:

$$\sum_{\mathbf{y}} \ell(\mathbf{y}, \hat{\mathbf{y}}) P(\mathbf{y}|\mathbf{x}) \approx \frac{1}{K} \sum_{i=1}^{K} \frac{\sum_{i \in \mathcal{V}} P_i(y_i = k|\mathbf{x}) \mathbb{1}\{\hat{y}_i = k\}}{\sum_{i \in \mathcal{V}} \left(\mathbb{1}\{\hat{y}_i = k\} + P_i(y_i = k|\mathbf{x}) \mathbb{1}\{\hat{y}_i \neq k\}\right)} \tag{14}$$

where $K$ is the number of classes; [13] and [16] proposed heuristics to optimize for this approximate expected IOU. However, since we restrict our solution set to a small set of candidate solutions, we do not require such procedures.

## 6.1. Estimating masses for toy graphs

As a sanity check, we used our method for computing mass around solutions on small toy $N \times N$ grids. The unary and pairwise log-potentials are randomly set in the interval $(-\infty, 0]$. We also sample a random (small) radius value ranging from 1 to $\sqrt{N}$. Fig. 3a shows absolute error in log-mass estimation, averaged across 10 runs, for Bethe and uniform sampling in the Hamming ball. We can see that the Bethe approximation works very well, while sampling takes a large number of samples, even in such a small graph.

## 6.2. Interactive Binary Segmentation

We use the foreground-background setup from [1] – 100 images were from the PASCAL VOC 2010 dataset, and manually annotated with scribbles for objects of interest.

**CRF construction:** For every superpixel in the image, outputs of Transductive SVMs are used as node potentials, which along with constrast sensitive Potts edge potentials provide the binary CRF to be used for segmentation.

**Solution Proposals:** We run DivMBest on the constructed CRF to acquire a set of 30 diverse solutions per test image.

**Evaluation:** 50 of these images were used to train the tSVM parameters, and cross validation was done on the remaining 50 images. Since this is a small dataset, we perform Leave-One-Out cross validation. The cross-validation is on the performance at $M = 30$.

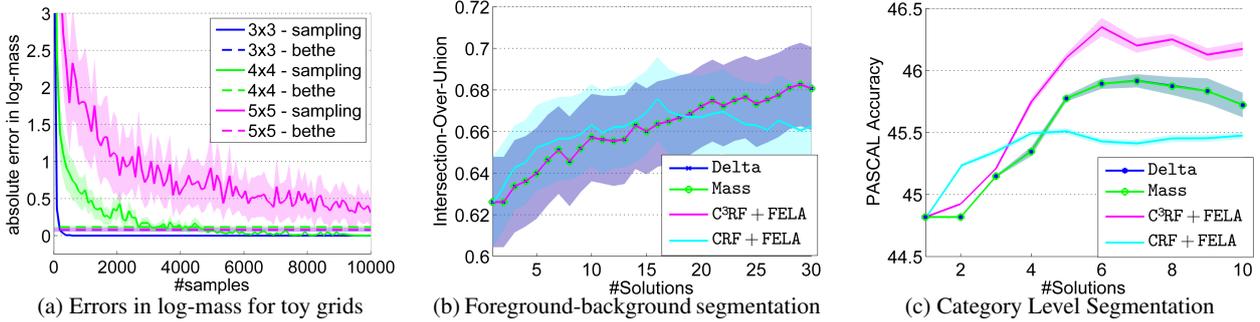

(a) Errors in log-mass for toy grids  (b) Foreground-background segmentation  (c) Category Level Segmentation

Figure 3: (a) Errors for the grids are approximately the same for the bethe approximation. Sampling requires more samples with increasing grid size to provide a good estimate. (b) All EMBR predictors improve significantly upon MAP. `Mass` and `C³RF+FELA` opt for a radius of 0 and reduce to `Delta`. CRF+FELA performs relatively poorly. (c) `C³RF+FELA` performs the best, with a $\sim 1.4\%$ improvement over MAP. Again, `Mass` performs similarly as `Delta`, and CRF+FELA performs relatively poorly.

The loss function is 1 minus intersection-over-union of the ground truth and predicted foreground superpixels.

**Results:** In Fig. 3b, we show how the various predictors perform. The shading around the curves indicate error bars. We observe that `Delta`, `Mass`, and `C³RF+FELA` perform exactly the same: 5.45% above MAP. This is because `Mass` and `C³RF+FELA` both pick a Hamming radius of 0 during cross-validation, and reduce to `Delta`. CRF+FELA appears to perform well at the start, however, at $M = 30$, it performs the poorest: 3.63% above MAP.

### 6.3. PASCAL Category Segmentation

We evaluate our predictors on the task of category segmentation - label each pixel with one of a set of categories - on the PASCAL VOC 2012 `val` set.

**CRF Construction:** Multi-label pairwise CRFs are constructed on superpixels of images from PASCAL VOC 2012 `train` and `val`. The node potentials are outputs of category-specific regressors, which are trained on `train` using [3]. Edge potentials are multi-label Potts.

**Solution Proposals:** We use solutions generated as in [14] - CPMC segments [4] scored by Support Vector Regressors over second-order pooled features [3] are greedily pasted. These solutions have a higher MAP accuracy of 44.82%, and also exhibit sufficient diversity. However, these solutions have many duplicates, so we pick the first 10 unique solutions from a set of 50 DivMBest solutions. In the few cases where there weren't 10 unique solutions among the 50 diverse solutions, we allowed duplicates to remain, the intuition being that since these solutions are repeatedly picked by the DivMBest algorithm, they represent exceptionally strong beliefs of the model, and allowing these duplicates to remain would assign them a larger weight.

**Evaluation:** The standard PASCAL evaluation criteria is the corpus-level Jaccard index, averaged over all 21 categories. Since we are performing instance level predictions, and then evaluating at a corpus level, the loss function used in all predictors is instance level loss, but the parameters are chosen by cross-validation on corpus-level loss.

We perform 10 fold cross-validation on `val`. To acquire a clearer sense of variance, we perform the 10 fold cross-val for 5 different random permutations of the corpus.

**Results:** Fig. 3c compares performances of the various predictors. The shading around the curves indicate error bars. We observe, yet again, that `Delta` and `Mass` perform very similarly - 0.9% above MAP.

CRF+FELA is once again relatively promising at the start, but performs poorest at $M = 10$: 0.68% above MAP. `C³RF+FELA` performs the best, with a 1.35% improvement over MAP, and 0.45% improvement over `Delta`.

### 6.4. Discussion

We note that `Mass` is not particularly helpful. In fact, cross-validation consistently picks `Delta` over `Mass` almost all of the time (by opting for a Hamming radius of 0). We hypothesize that this is because `Mass` makes a crude assumption that worsens the approximation quality with increasing radii. It is easy to see that the `Mass` predictor essentially assumed that the loss of all solutions in a Hamming ball are the same. Since we are labeling superpixels, flipping even a small number of superpixels typically results in losses that are significantly different from the solution we form the Hamming ball around, especially since we evaluate loss on a pixel level.

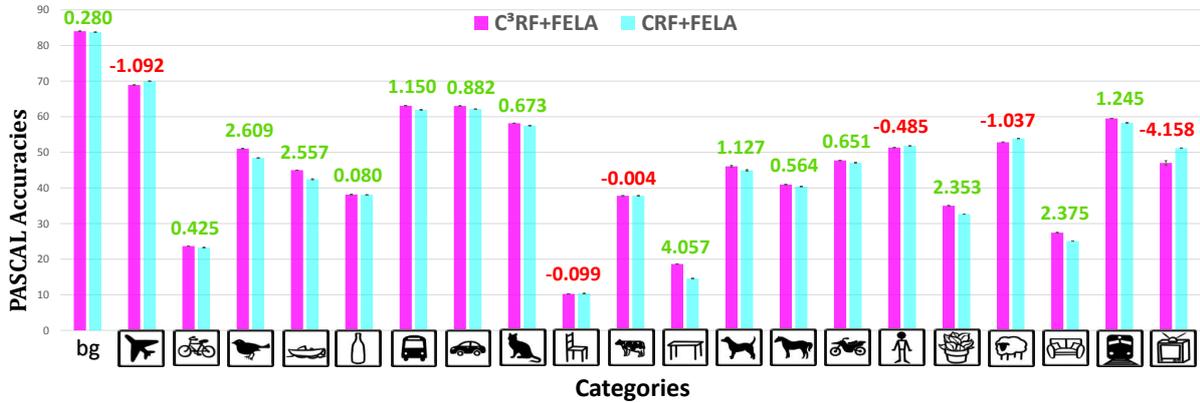

Figure 4: Comparing classwise accuracies for C³RF+FELA and CRF+FELA. The numbers on top of the bars indicate the difference in accuracy between C³RF+FELA and CRF+FELA.

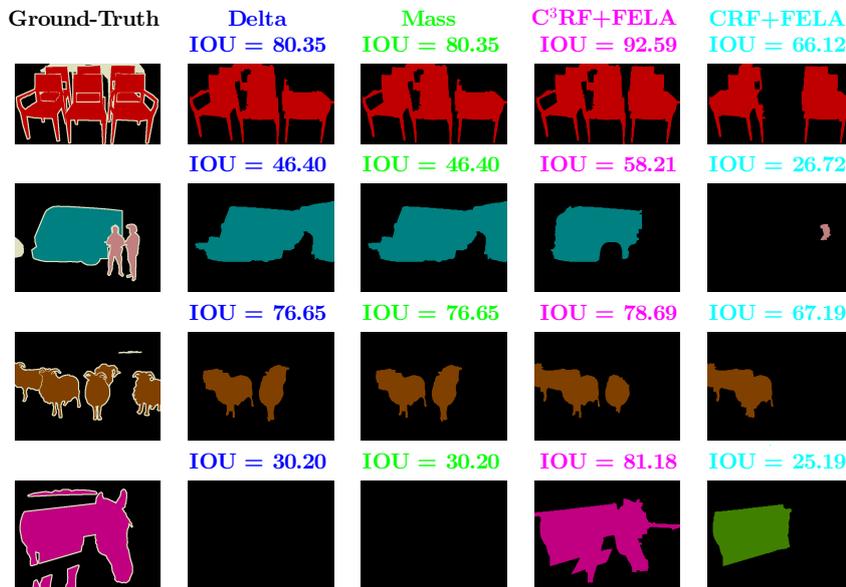

Figure 5: Qualitative examples showing the solutions predicted by the predictors for some cases where C³RF+FELA outperformed the other predictors. Since these are instance level accuracies, this is not a very accurate portrayal of performance at a corpus level.

**To bin or not to bin?** To look closer at the gains we achieve from the C³RF+FELA predictor, we take a look at the classwise PASCAL accuracies for C³RF+FELA and CRF+FELA.

In Fig. 4 we see that C³RF+FELA performs significantly better (or similarly) than CRF+FELA for most of the classes. For 4 of the classes - *aeroplane*, *person*, *sheep*, and *tvmonitor* - CRF+FELA outperforms C³RF+FELA.

In Fig. 6a we observe that, in general, the log-probability of the classes are higher under the C³RF marginals (which are just the mass-averaged marginals of all the Hamming constrained CRFs) than the marginals over the entire solutions space. This implies that the C³RF marginals might be better calibrated to minimize expected loss over the reduced distribution we are working with than the marginals from the non-constrained CRF.

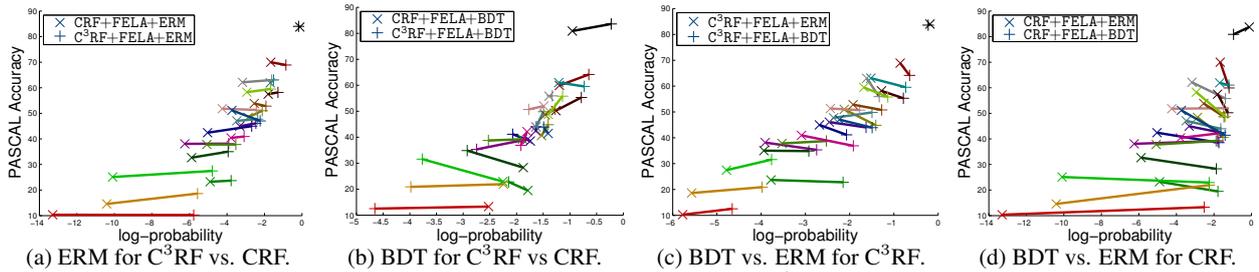

(a) ERM for C$^3$RF vs. CRF.  (b) BDT for C$^3$RF vs CRF.  (c) BDT vs. ERM for C$^3$RF.  (d) BDT vs. ERM for CRF.

Figure 6: BDT vs. ERM on PASCAL accuracy on C$^3$RF vs. CRF model.

(Normalized) log-probabilities were computed from pixel-wise marginals for every class $k$ as follows

$$\frac{1}{|GT(k)|} \sum_{i \in GT(k)} \log P_i(y_i = k). \tag{15}$$

where $GT(k)$ is the set of all pixels in the ground truth set of images labeled $k$.

**ERM or BDT?** Since the plot suggests that optimizing for log-probability might possibly lead to corresponding improvements in PASCAL accuracy, we also optimized our set of parameters for log-probability for the C$^3$RF and CRF models, thus performing approximate MBR in a BDT setting. We observed that this did not result in overall improved performance, although the performance of some classes improved when optimizing for log-probability. Interestingly, as shown in Fig. 6b, the overall performance for the C$^3$RF model is still higher when optimizing for log-probability. This could be further indication of the benefits of a candidate-contrained model when performing EMBR.

We also look at similar plots (Fig. 6c and Fig. 6d) comparing classwise performance when optimizing for log-probability vs. optimizing for PASCAL accuracy for the C$^3$RF and CRF models separately. It would appear that for both cases, ERM results in overall better performance than maximum likelihood training.

Table 1 shows the overall PASCAL accuracies for the 4 experiments.

|  | MAP | ERM | BDT |
|---|---|---|---|
| C$^3$RF+FELA | 44.82 | 46.17 | 44.44 |
| CRF+FELA | 44.82 | 45.50 | 42.42 |

Table 1: Overall PASCAL accuracies for C$^3$RF vs. CRF model when training parameters via ERM vs. BDT

We believe that since our models are not well-specified in the first place, MLE of the predictor parameters is not a good choice, as our experiments demonstrate, and that ERM is more suitable.

# 7. Conclusion

This work develops a methodology for loss-aware structured prediction that is compatible with the pipeline systems that dominate computer vision benchmarks today. The key idea is to make use of the pipeline systems to generate high quality candidate solutions, and use the candidate solutions to focus the probability mass of a CRF, yielding a Candidate Constrained CRF. Our focus in this work was on using C$^3$RF for loss-aware prediction for loss functions that admit accurate factorized approximations, but we believe the approach to potentially be more general. In future work we would like to explore applying similar ideas to more general classes of loss functions, and perhaps even beyond loss-aware prediction.

# Supplementary Material for
# Candidate Constrained CRFs for Loss-Aware Structured Prediction

In this supplement we investigate why the predictor `Mass` does not improve over `Delta` (Section 1 and Section 2). We also visualize the marginals from the C$^3$RF and the unconstrained CRF (Section 3).

## 1. Correlation between masses and scores of candidate solutions

Note that if the masses of the candidate solutions are similar to the scores, the masses provide no additional information than the scores, and `Mass` would not perform significantly different from `Delta`. To inspect how relatively different masses are from scores, we compute rank correlations between masses and scores of candidate solutions.

In Fig. 1, we look at histograms of rank correlations between masses and scores for three different temperatures and increasing radii for the 1449 images in PASCAL VOC 2012 `val`. We note that for smaller radii, there is near perfect positive correlation between masses and scores. This indicates that the predictor `Mass` does not utilize any more information than `Delta`. At larger radii, the correlation drops a little, but as we show next, with larger radii, there is a progressively cruder approximation of the expected loss. We find, empirically, that larger radii is not chosen by cross-validation. Together, these two observations explain why `Mass` performs identical to `Delta` in our experiments.

## 2. `Mass` assumes constant losses in Hamming balls

We now show that the predictor `Mass` assumes that losses of all solutions in a Hamming ball are the same as the loss of the candidate solution.

Recall that we are minimizing expected loss over a Hamming constrained distribution

$$\mathbf{y}_{\mathbf{Y},R}^{\text{C}^3\text{RF+FELA}} = \operatorname*{argmin}_{\hat{\mathbf{y}} \in \mathbf{Y}} \sum_{\mathbf{c} \in \mathbf{Y}} \sum_{\mathbf{y}' \in \mathcal{Y}} \mathcal{Z}(\{\mathbf{c}\}, R) P_{\{\mathbf{c}\},R}(\mathbf{y}'|\mathbf{x}) \ell(\mathbf{y}', \hat{\mathbf{y}}) \tag{1}$$

If we assume that the loss of all candidate solutions in a Hamming ball is the same, and it is the loss of the candidate solutions, Eq. 1 changes to `Mass`:

$$\begin{aligned}
\mathbf{y}_{\mathbf{Y},R}^{\text{C}^3\text{RF+FELA}} &= \operatorname*{argmin}_{\hat{\mathbf{y}} \in \mathbf{Y}} \sum_{\mathbf{c} \in \mathbf{Y}} \sum_{\mathbf{y}' \in \mathcal{Y}} \mathcal{Z}(\{\mathbf{c}\}, R) P_{\{\mathbf{c}\},R}(\mathbf{y}'|\mathbf{x}) \ell(\mathbf{y}', \hat{\mathbf{y}}) \\
&\approx \operatorname*{argmin}_{\hat{\mathbf{y}} \in \mathbf{Y}} \sum_{\mathbf{c} \in \mathbf{Y}} \ell(\mathbf{c}, \hat{\mathbf{y}}) \mathcal{Z}(\{\mathbf{c}\}, R) \sum_{\mathbf{y}' \in \mathcal{Y}} P_{\{\mathbf{c}\},R}(\mathbf{y}'|\mathbf{x}) \\
&= \operatorname*{argmin}_{\hat{\mathbf{y}} \in \mathbf{Y}} \sum_{\mathbf{c} \in \mathbf{Y}} \ell(\mathbf{c}, \hat{\mathbf{y}}) \mathcal{Z}(\{\mathbf{c}\}, R) = \mathbf{y}_{\mathbf{Y},R}^{\text{Mass}}
\end{aligned} \tag{2}$$

Clearly, the assumption that "the loss of all solutions within a large Hamming ball can be represented by the candidate solution" gets more inaccurate with an increasing Hamming radius. Hence, the only recourse is to opt for a smaller radius, which unfortunately provides the same information as the scores themselves.

## 3. Visualizing marginals

In Fig. 3, Fig. 4, Fig. 5, Fig. 6, and Fig. 7, we visualize marginals for different radii and temperature for the first five images from PASCAL VOC 2012 `val`.

For all figures, the top row corresponds to marginals with $R = 0\%$ of number of nodes, which is essentially a mixture of the candidate solutions weighted by the scores of the solutions. The bottom row corresponds to marginals from $R = 100\%$ of number of nodes, which is just the marginals on the unconstrained CRF. The middle rows are intermediate values of $R$, which are the mass-averaged marginals from the Hamming-constrained CRFs.

Visually, we can observe that higher temperatures lead to more uniform marginals, and larger radii lead to marginals with more "even" support over the set of all configurations.

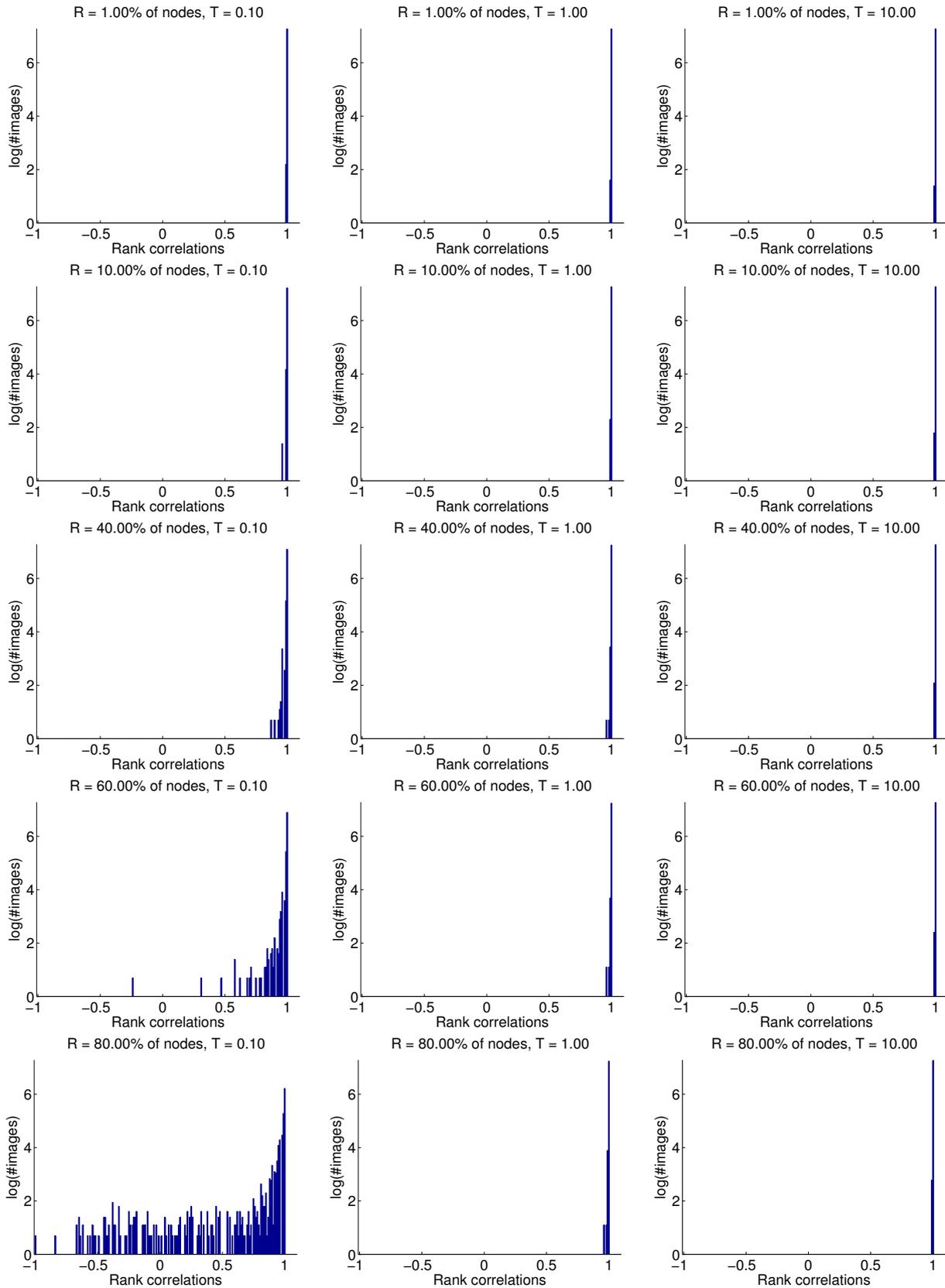

Figure 1: Histograms of rank correlations between masses of solutions and their scores. The high positive correlation at smaller radii indicate that `Mass` and `Delta` would make the same predictions.

| aeroplane | bicycle | bird | boat | bottle |
| bus | car | cat | chair | cow |
| diningtable | dog | horse | motorbike | person |
| pottedplant | sheep | sofa | train | tvmonitor |
| background | | | | |

Figure 2: Color code for PASCAL classes.

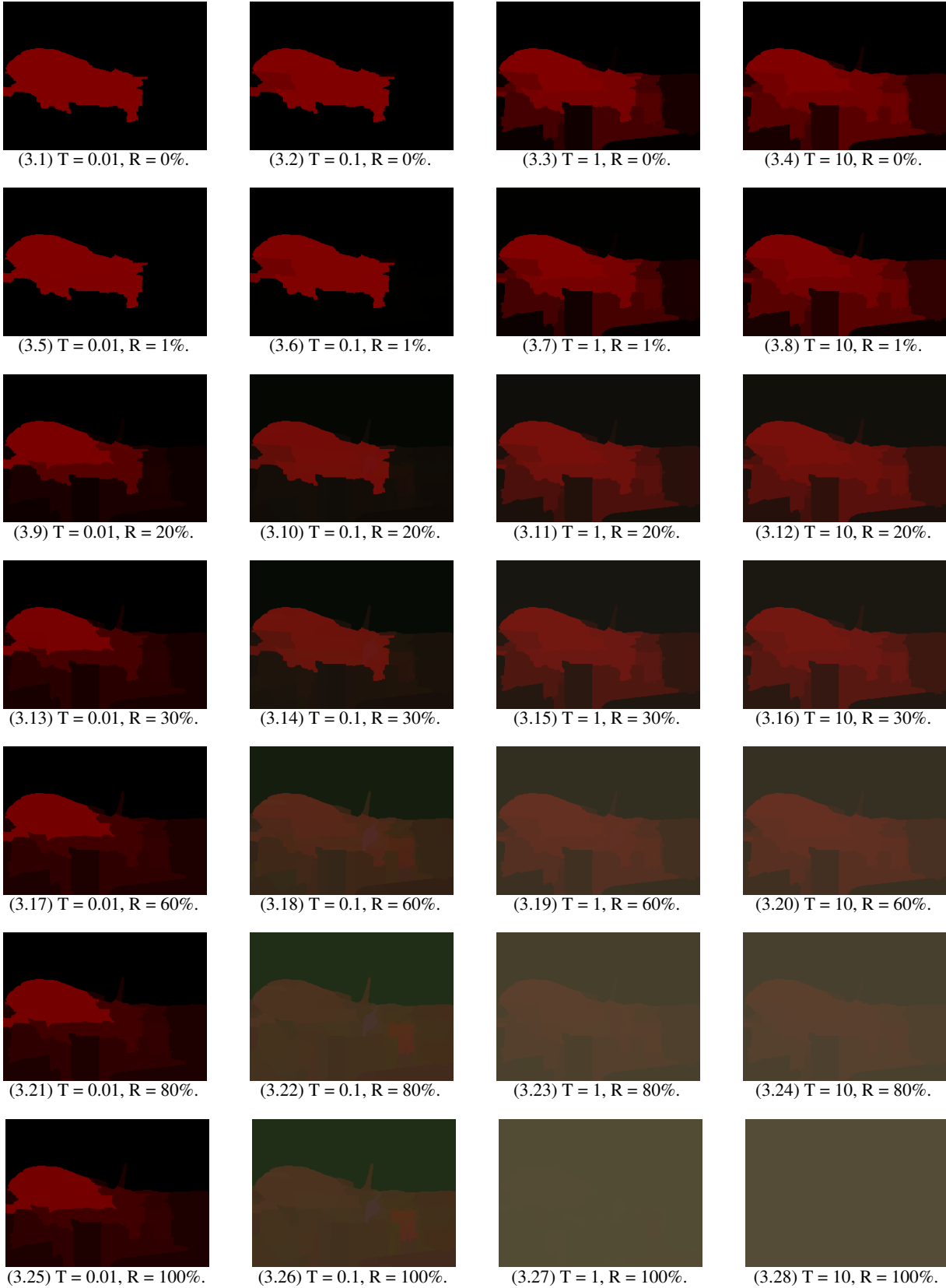

Figure 3: Image 1 from `val`: Top row corresponds to marginals with $R = 0\%$ of number of nodes, bottom row corresponds to marginals from $R = 100\%$ of number of nodes, and middle rows are intermediate values of $R$.

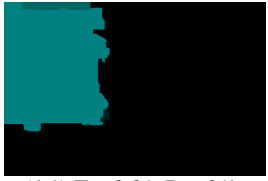 (4.1) T = 0.01, R = 0%.
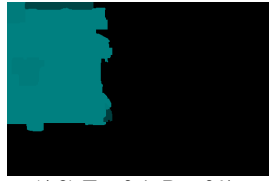 (4.2) T = 0.1, R = 0%.
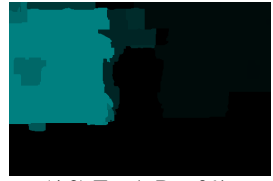 (4.3) T = 1, R = 0%.
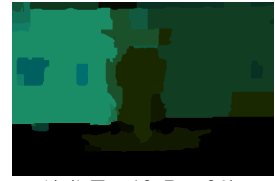 (4.4) T = 10, R = 0%.
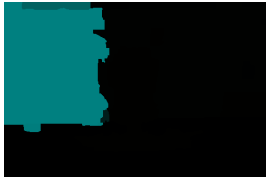 (4.5) T = 0.01, R = 1%.
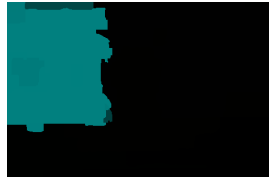 (4.6) T = 0.1, R = 1%.
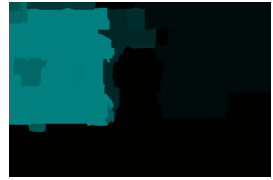 (4.7) T = 1, R = 1%.
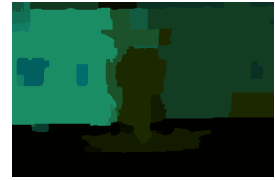 (4.8) T = 10, R = 1%.
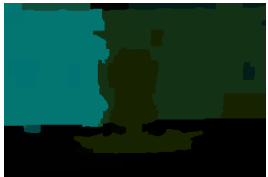 (4.9) T = 0.01, R = 20%.
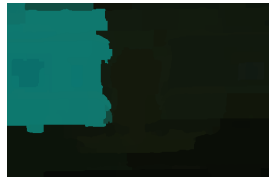 (4.10) T = 0.1, R = 20%.
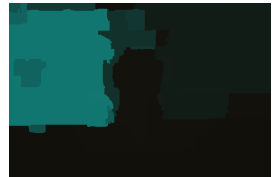 (4.11) T = 1, R = 20%.
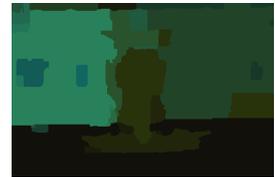 (4.12) T = 10, R = 20%.
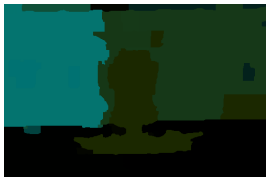 (4.13) T = 0.01, R = 30%.
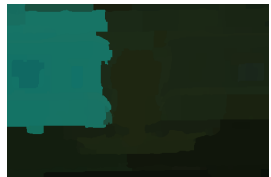 (4.14) T = 0.1, R = 30%.
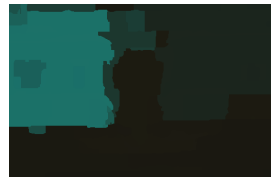 (4.15) T = 1, R = 30%.
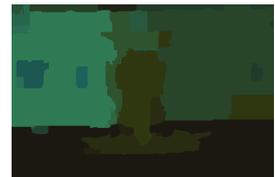 (4.16) T = 10, R = 30%.
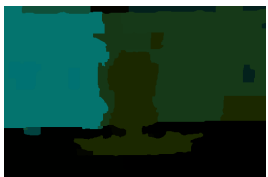 (4.17) T = 0.01, R = 60%.
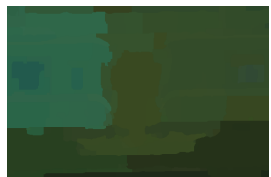 (4.18) T = 0.1, R = 60%.
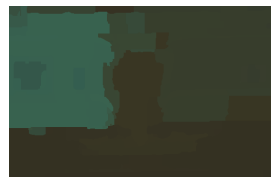 (4.19) T = 1, R = 60%.
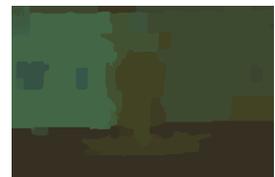 (4.20) T = 10, R = 60%.
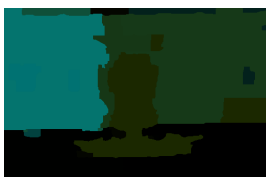 (4.21) T = 0.01, R = 80%.
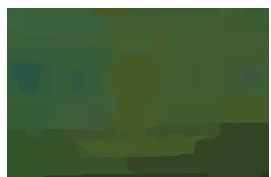 (4.22) T = 0.1, R = 80%.
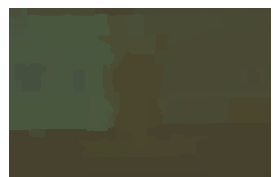 (4.23) T = 1, R = 80%.
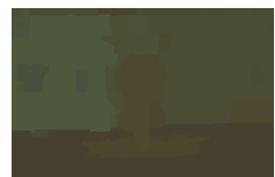 (4.24) T = 10, R = 80%.
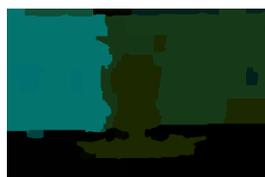 (4.25) T = 0.01, R = 100%.
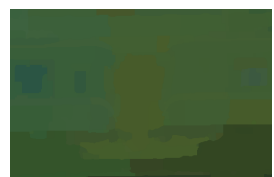 (4.26) T = 0.1, R = 100%.
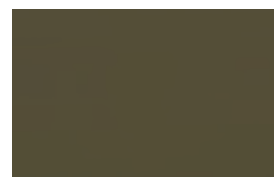 (4.27) T = 1, R = 100%.
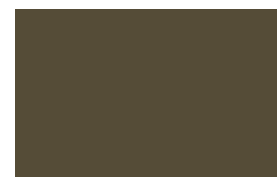 (4.28) T = 10, R = 100%.

Figure 4: Image 2 from `val`: Top row corresponds to marginals with $R = 0\%$ of number of nodes, bottom row corresponds to marginals from $R = 100\%$ of number of nodes, and middle rows are intermediate values of $R$.

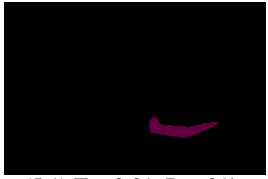 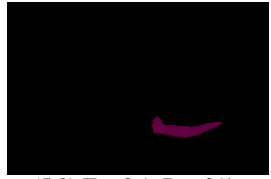 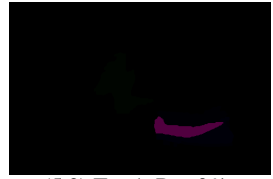 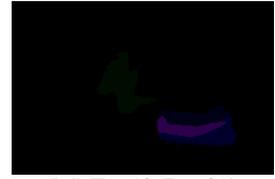

(5.1) T = 0.01, R = 0%.    (5.2) T = 0.1, R = 0%.    (5.3) T = 1, R = 0%.    (5.4) T = 10, R = 0%.

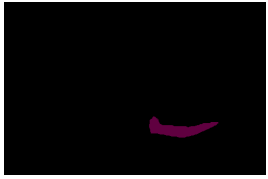 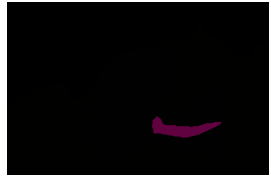 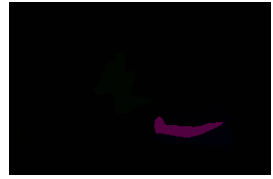 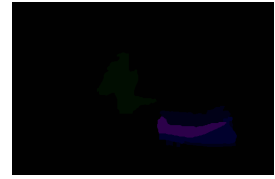

(5.5) T = 0.01, R = 1%.    (5.6) T = 0.1, R = 1%.    (5.7) T = 1, R = 1%.    (5.8) T = 10, R = 1%.

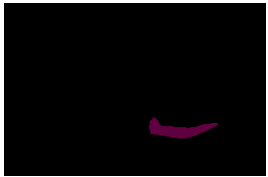 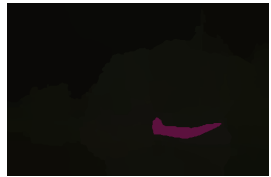 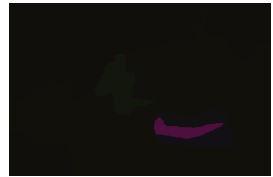 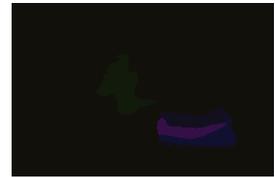

(5.9) T = 0.01, R = 20%.    (5.10) T = 0.1, R = 20%.    (5.11) T = 1, R = 20%.    (5.12) T = 10, R = 20%.

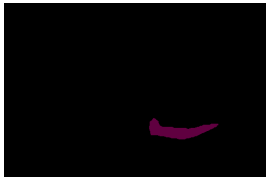 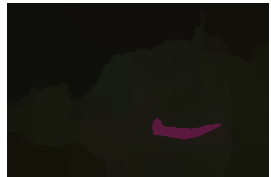 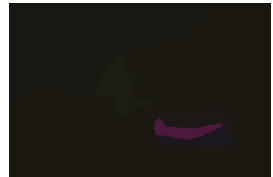 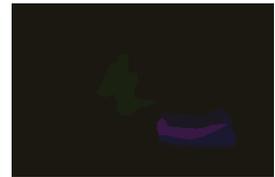

(5.13) T = 0.01, R = 30%.    (5.14) T = 0.1, R = 30%.    (5.15) T = 1, R = 30%.    (5.16) T = 10, R = 30%.

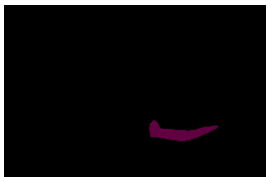 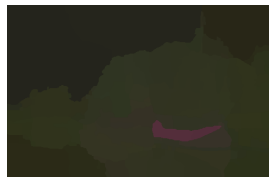 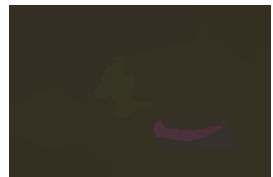 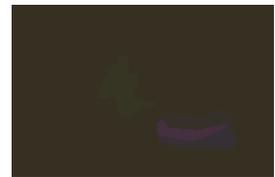

(5.17) T = 0.01, R = 60%.    (5.18) T = 0.1, R = 60%.    (5.19) T = 1, R = 60%.    (5.20) T = 10, R = 60%.

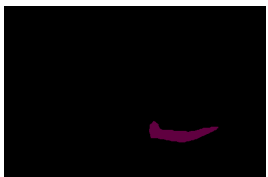 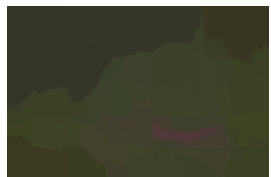 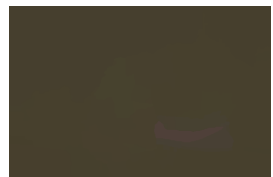 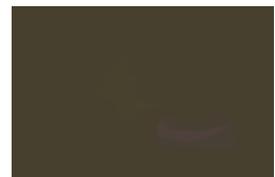

(5.21) T = 0.01, R = 80%.    (5.22) T = 0.1, R = 80%.    (5.23) T = 1, R = 80%.    (5.24) T = 10, R = 80%.

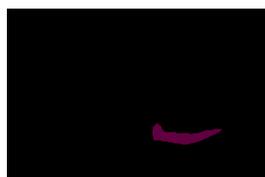 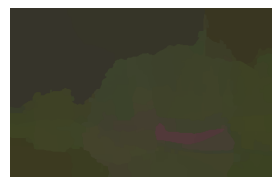 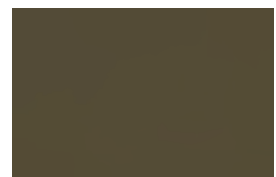 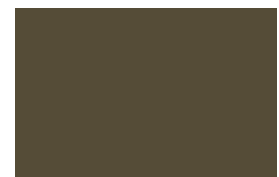

(5.25) T = 0.01, R = 100%.    (5.26) T = 0.1, R = 100%.    (5.27) T = 1, R = 100%.    (5.28) T = 10, R = 100%.

Figure 5: Image 3 from `val`: Top row corresponds to marginals with $R = 0\%$ of number of nodes, bottom row corresponds to marginals from $R = 100\%$ of number of nodes, and middle rows are intermediate values of $R$.

(6.1) T = 0.01, R = 0%.    (6.2) T = 0.1, R = 0%.    (6.3) T = 1, R = 0%.    (6.4) T = 10, R = 0%.

(6.5) T = 0.01, R = 1%.    (6.6) T = 0.1, R = 1%.    (6.7) T = 1, R = 1%.    (6.8) T = 10, R = 1%.

(6.9) T = 0.01, R = 20%.    (6.10) T = 0.1, R = 20%.    (6.11) T = 1, R = 20%.    (6.12) T = 10, R = 20%.

(6.13) T = 0.01, R = 30%.    (6.14) T = 0.1, R = 30%.    (6.15) T = 1, R = 30%.    (6.16) T = 10, R = 30%.

(6.17) T = 0.01, R = 60%.    (6.18) T = 0.1, R = 60%.    (6.19) T = 1, R = 60%.    (6.20) T = 10, R = 60%.

(6.21) T = 0.01, R = 80%.    (6.22) T = 0.1, R = 80%.    (6.23) T = 1, R = 80%.    (6.24) T = 10, R = 80%.

(6.25) T = 0.01, R = 100%.    (6.26) T = 0.1, R = 100%.    (6.27) T = 1, R = 100%.    (6.28) T = 10, R = 100%.

Figure 6: Image 4 from `val`: Top row corresponds to marginals with $R = 0\%$ of number of nodes, bottom row corresponds to marginals from $R = 100\%$ of number of nodes, and middle rows are intermediate values of $R$.

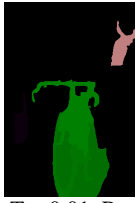 (7.1) T = 0.01, R = 0%.
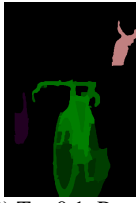 (7.2) T = 0.1, R = 0%.
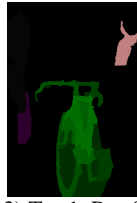 (7.3) T = 1, R = 0%.
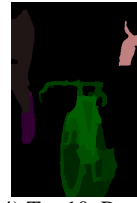 (7.4) T = 10, R = 0%.

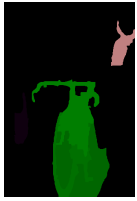 (7.5) T = 0.01, R = 1%.
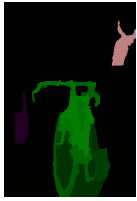 (7.6) T = 0.1, R = 1%.
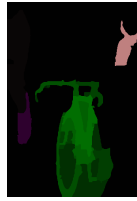 (7.7) T = 1, R = 1%.
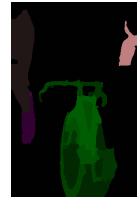 (7.8) T = 10, R = 1%.

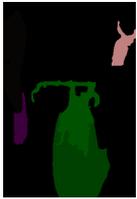 (7.9) T = 0.01, R = 20%.
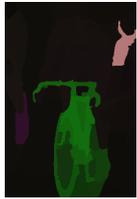 (7.10) T = 0.1, R = 20%.
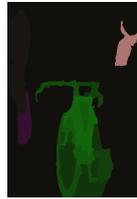 (7.11) T = 1, R = 20%.
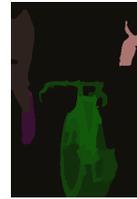 (7.12) T = 10, R = 20%.

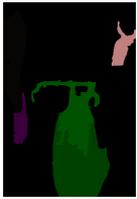 (7.13) T = 0.01, R = 30%.
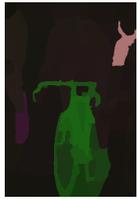 (7.14) T = 0.1, R = 30%.
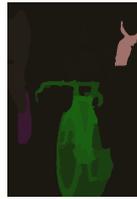 (7.15) T = 1, R = 30%.
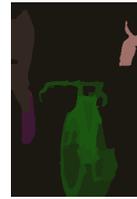 (7.16) T = 10, R = 30%.

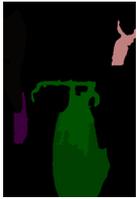 (7.17) T = 0.01, R = 60%.
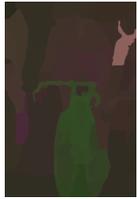 (7.18) T = 0.1, R = 60%.
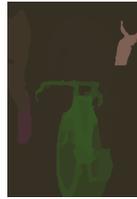 (7.19) T = 1, R = 60%.
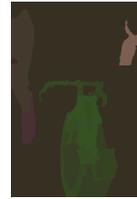 (7.20) T = 10, R = 60%.

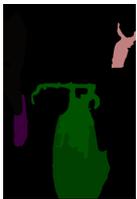 (7.21) T = 0.01, R = 80%.
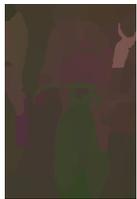 (7.22) T = 0.1, R = 80%.
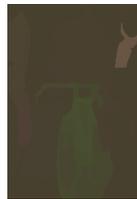 (7.23) T = 1, R = 80%.
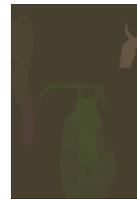 (7.24) T = 10, R = 80%.

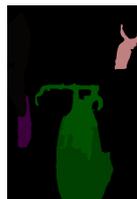 (7.25) T = 0.01, R = 100%.
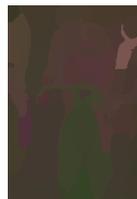 (7.26) T = 0.1, R = 100%.
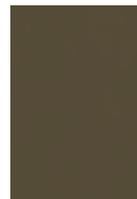 (7.27) T = 1, R = 100%.
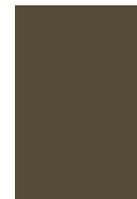 (7.28) T = 10, R = 100%.

Figure 7: Image 5 from `val`: Top row corresponds to marginals with $R = 0\%$ of number of nodes, bottom row corresponds to marginals from $R = 100\%$ of number of nodes, and middle rows are intermediate values of $R$.